\documentclass[journal]{journal}
\usepackage{graphicx,color,multicol,url}
\begin{document}
%
\title{Challenges in video based object detection in maritime scenario using computer vision}
\author{
\IEEEauthorblockN{Dilip~K.~Prasad \IEEEauthorrefmark{1}\IEEEauthorrefmark{6}, C. Krishna Prasath\IEEEauthorrefmark{1}, Deepu Rajan\IEEEauthorrefmark{2}, Lily Rachmawati\IEEEauthorrefmark{3}, Eshan Rajabally\IEEEauthorrefmark{4} and Chai Quek\IEEEauthorrefmark{2}}\\
\IEEEauthorblockA{\IEEEauthorrefmark{1}Rolls-Royce@NTU Corporate Lab, Singapore\\
}
\and
\IEEEauthorblockA{\IEEEauthorrefmark{2}School of Computer Science and Engineering,
Nanyang Technological University, Singapore\\}
\and
\IEEEauthorblockA{\IEEEauthorrefmark{3}Rolls-Royce plc, Singapore\\}
\and
\IEEEauthorblockA{\IEEEauthorrefmark{4}Rolls-Royce Derby, United Kingdom\\
\IEEEauthorrefmark{6}Email: dilipprasad@gmail.com}
}

\maketitle
\thispagestyle{empty}

\begin{abstract}
This paper discusses the technical challenges in maritime image processing and machine vision problems for video streams generated by cameras. Even well documented problems of horizon detection and registration of frames in a video are very challenging in maritime scenarios. More advanced problems of background subtraction and object detection in video streams are very challenging. Challenges arising from the dynamic nature of the background, unavailability of static cues, presence of small objects at distant backgrounds, illumination effects, all contribute to the challenges as discussed here.
\end{abstract}

\IEEEpeerreviewmaketitle

\section{Introduction}\label{sec:introduction}

\IEEEPARstart{W}{hile} computer vision techniques have advanced video processing and intelligence generation for several challenging dynamic scenarios, research in computer vision for maritime is still in nascent state and several challenges remain open in this field \cite{prasad2017survey}. This paper presents some of the challenges unique to the maritime domain.

\begin{figure}
  \centering
  \includegraphics[width=\linewidth]{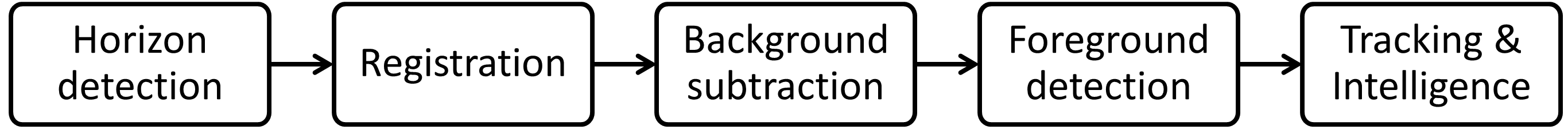}\\
  \vspace{-3mm}
  \caption{Simple block diagram for maritime video processing. }\label{fig:SimpleBlock}
  \vspace{-3mm}
\end{figure}

A simple block diagram for processing of maritime videos is given in Fig. \ref{fig:SimpleBlock}, where the objective is to track foreground objects and generate intelligence and situation-awareness. Foreground objects are the objects anchored, floating, or navigating in water, including sea vessels, small personal boats and kayaks, buoys, debris, etc. Air vehicles, birds, and fixed structures, such as in ports, qualify as outliers or background. Also, wakes, foams, clouds, water speckle, etc. qualify as background. The first four blocks form the core of video processing and the performance of these blocks directly affect the attainment of the objective. The challenges specific to these four blocks are discussed in the sections \ref{sec:horizon} to \ref{sec:foreground}, respectively. The challenges due to weather are discussed in section \ref{sec:weather}.

We use 3 datasets from three different sources to illustrate the challenges. Two datasets are from the external sources, buoy dataset \cite{fefilatyev2012detection} and Mar-DCT dataset \cite{bloisi2015}. The camera in the buoy dataset is mounted on a floating buoy which is subject to significant amount of motion from one frame to another. The camera used in Mar-DCT dataset is mounted on a stationary platform on-shore. Sometimes, zoom operations are used while capturing the videos. The third dataset Singapore-Marine-dataset is created by the authors using Canon 70D camera. Videos are acquired in two scenarios, namely at sea (videos captured on-board a vessel in motion) and on-shore (videos captured with camera on a stationary platform on-shore). The details of the datasets are presented in Table \ref{tab:Datasets}.

\section{Horizon detection}\label{sec:horizon}

\begin{figure}
  \centering
  \includegraphics[width=0.6\linewidth]{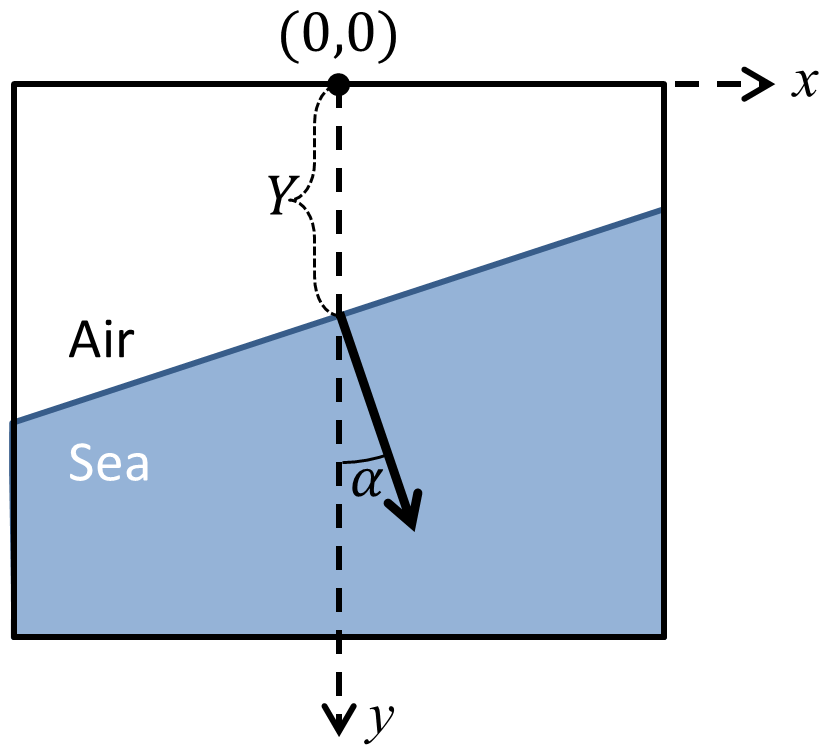}
  \vspace{-3mm}
  \caption{Representation of horizon using $Y$ and $\alpha$.\vspace{-3mm}}\label{fig:horizonRep}
\end{figure}

We represent horizon using two parameters, the vertical position $Y$ of the center of the horizon from the upper edge of the image, and the angular position $\alpha$ made by the horizon with the horizontal axis. This is illustrated in Fig. \ref{fig:horizonRep}. In the case of cameras mounted on mobile platform, the vertical and angular position is subject to large amount of motion, as noted in Table \ref{tab:Datasets}. In Table \ref{tab:Datasets}, E($Y$) and E($\alpha$) represent the mean values of $Y$ and $\alpha$ for a video. The ground truth for horizon is generated for each frame of these videos manually using independent volunteers \cite{prasad2016Tencon}.

\begin{table}[!b]
\vspace{-3mm}
\caption{Details of the datasets used in this paper.}\label{tab:Datasets}
  \centering
  \vspace{-3mm}
  \begin{tabular}{|l|r|r|r|r|}
    \hline
    {Camera} & \multicolumn{2}{c|}{At sea} & \multicolumn{2}{c|}{On-shore}\\
    \hline
    \hline
    {Datasets}& Buoy & \multicolumn{2}{c|}{Singapore-Marine-Dataset} & Mar-DCT \\
    \hline
    \hline
    Number & 10   & 11        & 28    & 9 \\
    of videos & & & & \\
    \hline
    Number & 998     & 2772      & 12604 & 7410 \\
    of frames & & & & \\
    \hline
    \hline
    \multicolumn{5}{|c|}{Horizon related}\\
    \hline
    min($Y$-{\rm E}($Y$)) & -281.68 & -436.30 & -13.54 & -52.32\\
    (pixels) & & & & \\
    \hline
    max($Y$-{\rm E}($Y$))& 307.82 & 467.86 & 9.95 & 35.69\\
    (pixels) & & & & \\\hline
    Std. dev. & 107.98    & 145.10    & 1.52  & 9.98 \\
    of $Y$ (pixels) & & & & \\
    \hline
    \hline
    min($\alpha$-{\rm E}($\alpha$)) & -15.72 & -26.34 & -0.99 & -1.25 \\
    (degree) & & & & \\
    \hline
    max($\alpha$-{\rm E}($\alpha$))& 20.72 & 12.99 & 0.51 & 1.75\\
    (degree) & & & & \\
    \hline
    Std. dev. & 4.40     & 1.11      & 0.04  & 0.22 \\
    of $\alpha$ (degree) & & & & \\
    \hline
    \hline
    \multicolumn{5}{|c|}{Objects related}\\
    \hline
    Min number & 0 & 0 & 0 & 1\\
    of objects &  & & & \\ \hline
    Max. number & 3 & 10 & 20 & 2\\
    of objects & & & & \\ \hline
  \end{tabular}
\end{table}

We discuss two state-of-the-art methods \cite{fefilatyev2012detection,ettinger2003vision}, which we succinctly refer to as FGSL (abbreviation derived from the first alphabets of the authors' names) \cite{fefilatyev2012detection} and ENIW (abbreviation derived from the first alphabets of the authors' names) \cite{ettinger2003vision}, in the context of the present datasets. They are chosen as they both use a combination of two main approaches used for horizon detection, as discussed next.

One popular approach is to detect the most prominent line feature through parametric projection of edges in the image space to the parametric space of line features, such as Hough transform (HT). This approach assumes that horizon appears as a long line feature in the image. We note that this approach uses projective mappings and parametric space and is different from another line of research on line fitting on edge maps \cite{prasad2012novel,prasad2012polygonal,prasad2014deb}. Although we do not exclude the utility of dominant point detection and line fitting \cite{prasad2013online,prasad2013fabrication} for horizon detection in the on-board maritime detection problems, we note that no research work on horizon detection has so far employed these techniques.

The second popular approach is to select a candidate horizon solution that maximizes the statistical distances between the color distributions \cite{cheng2014illuminant} of the two regions created by the candidate solution. This approach assumes that sea and sky regions have color distributions with large statistical distance between them and that the candidate solution separates the regions into sea and sky regions. While they are similar in using statistical distribution as the main criterion and using prominent linear features as candidate solutions, they are different in the choice of statistical distance measures.

\begin{table}[!b]
\vspace{-3mm}
  \caption{Statistics of errors in $Y$ for different methods are listed here.}\label{tab:horizonresults}
  \vspace{-3mm}
  \centering
  \begin{tabular}{|l|r|r|r|r|}
    \hline
     {} & Buoy & On-board & On-shore & Mar-DCT \\
    \hline
    \hline
    \multicolumn{5}{|c|}{Error in $Y$} \\
     \hline
    \hline
    {} & \multicolumn{4}{|c|}{25th percentile (1st quartile)} \\
     \hline
     ENIW   & 0.92    & 71.82     & 15.30     & 1.38  \\
     FGSL   & 0.72    & 72.06     & 7.30    & 4.29  \\
     \hline
     {} & \multicolumn{4}{|c|}{50th percentile (median)}\\
     \hline
     ENIW & 1.93     & 117.81     & 115.25     & 37.43 \\
     FGSL & 1.59     & 118.14     & 115.25     & 198.58 \\
     \hline
    \hline
     \multicolumn{5}{|c|}{Error in $\alpha$} \\
    \hline
    \hline
     {} & \multicolumn{4}{|c|}{25th percentile (1st quartile)} \\
     \hline
     ENIW   & 0.24     & 0.47     & 0.18     & 0.26  \\
     FGSL   & 0.20     & 0.49     & 0.18     & 0.64  \\
     \hline
     {} & \multicolumn{4}{|c|}{50th percentile (median)}\\
     \hline
     ENIW & 0.46     & 1.10     & 0.38     & 1.18 \\
     FGSL & 0.38     & 1.19     & 0.35     & 1.00  \\
    \hline
  \end{tabular}
\end{table}

\begin{figure*}[!t]
  \centering
  \includegraphics[width=\linewidth]{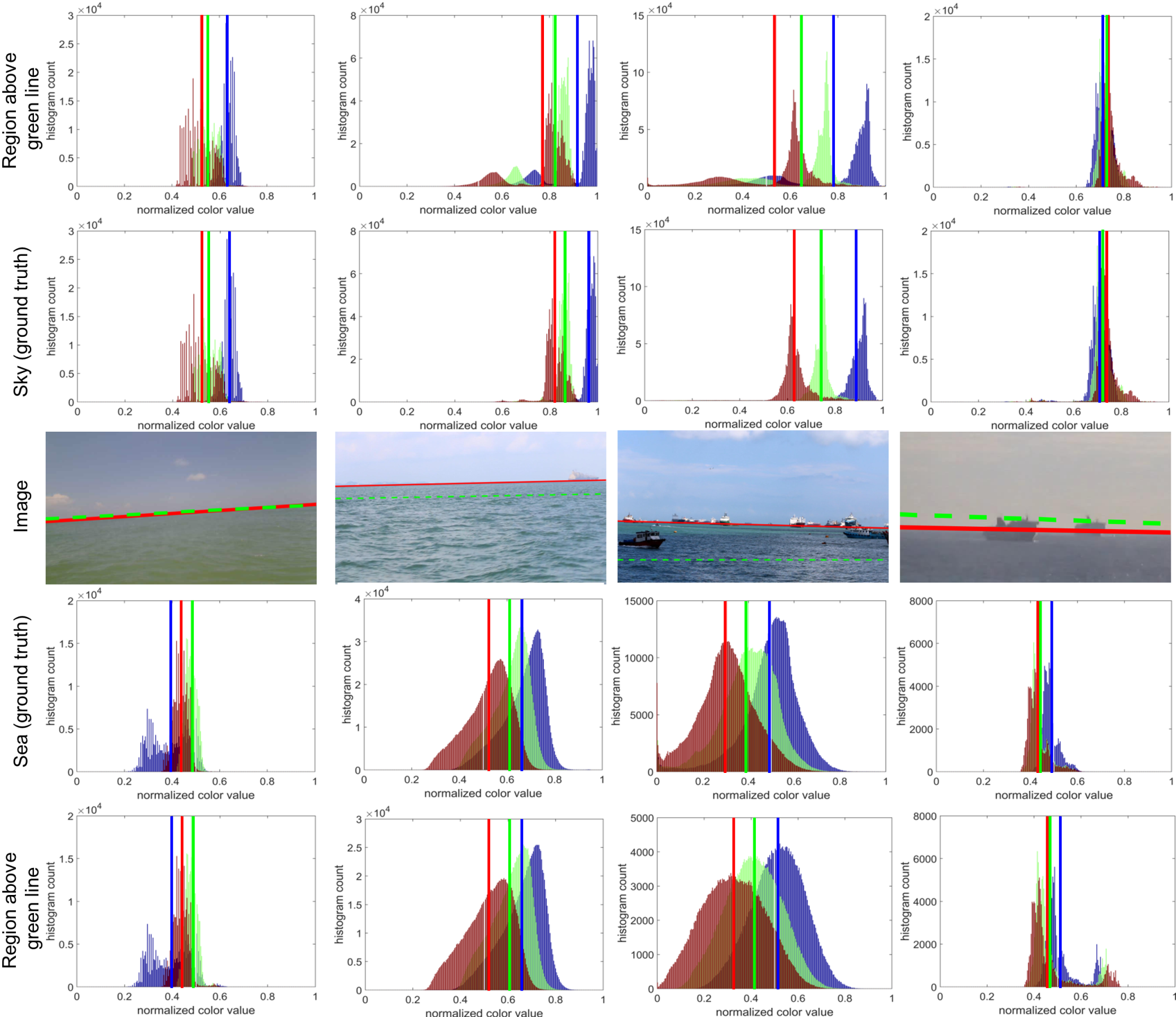}\\
  \vspace{-3mm}
  \caption{Statistical distribution of the sea and sky regions determined by the horizon ground truth (solid red line) and the upper and lower regions determined by the most prominent HT candidate (green dashed lines).\vspace{-3mm}}\label{fig:horizon}
\end{figure*}

The performance of these methods is presented in Table \ref{tab:horizonresults}. It is seen that the methods perform extremely well for Buoy dataset but perform poorly for the other datasets in terms of the vertical position of the horizon. In Fig. \ref{fig:horizon}, we show that the assumption behind the statistical approach used by both methods may not apply. We present one image from each dataset (3rd row), the horizon ground truth (red solid line), the most prominent HT candidate (green dashed line), and the color distributions of the regions created by them in Fig. \ref{fig:horizon}. For the first image, it is seen that the HT candidate for the horizon matches the ground truth and indeed the color distributions corresponding to the sea and sky regions match well. However, for the other three images, the Hough transform candidates do not match with the ground truth. Let us first consider the upper regions created by the ground truth and the Hough candidates. For the Singapore-Marine dataset, the upper region created by the Hough candidate includes the sky region and part of the sea region. This causes some change in the color distribution at lower color values. Nevertheless, the distribution is clearly dominated by sky and statistical distance metrics may not be effective in distinguishing sea and sky regions effectively. For example, the mean values (shown using vertical lines in the color distribution plots) of the distributions corresponding to the incorrect horizon are not significantly different from the mean values of the distributions corresponding to the ground truth. Numerically, the means show the same shift for all the color channels between the incorrect horizon and the ground truth. The maximum shift of 25 value (between 0 to 256 digital values) is observed for the third image for the sky region. The shift is caused by the inclusion of part of the sea in the upper region. In the other cases, the typical shift is 0 value to 5 values. The same observation applies to the example from Mar-DCT dataset as well, however with the shift observed in the bottom region.

\begin{figure}
  \centering
  \includegraphics[width=\linewidth]{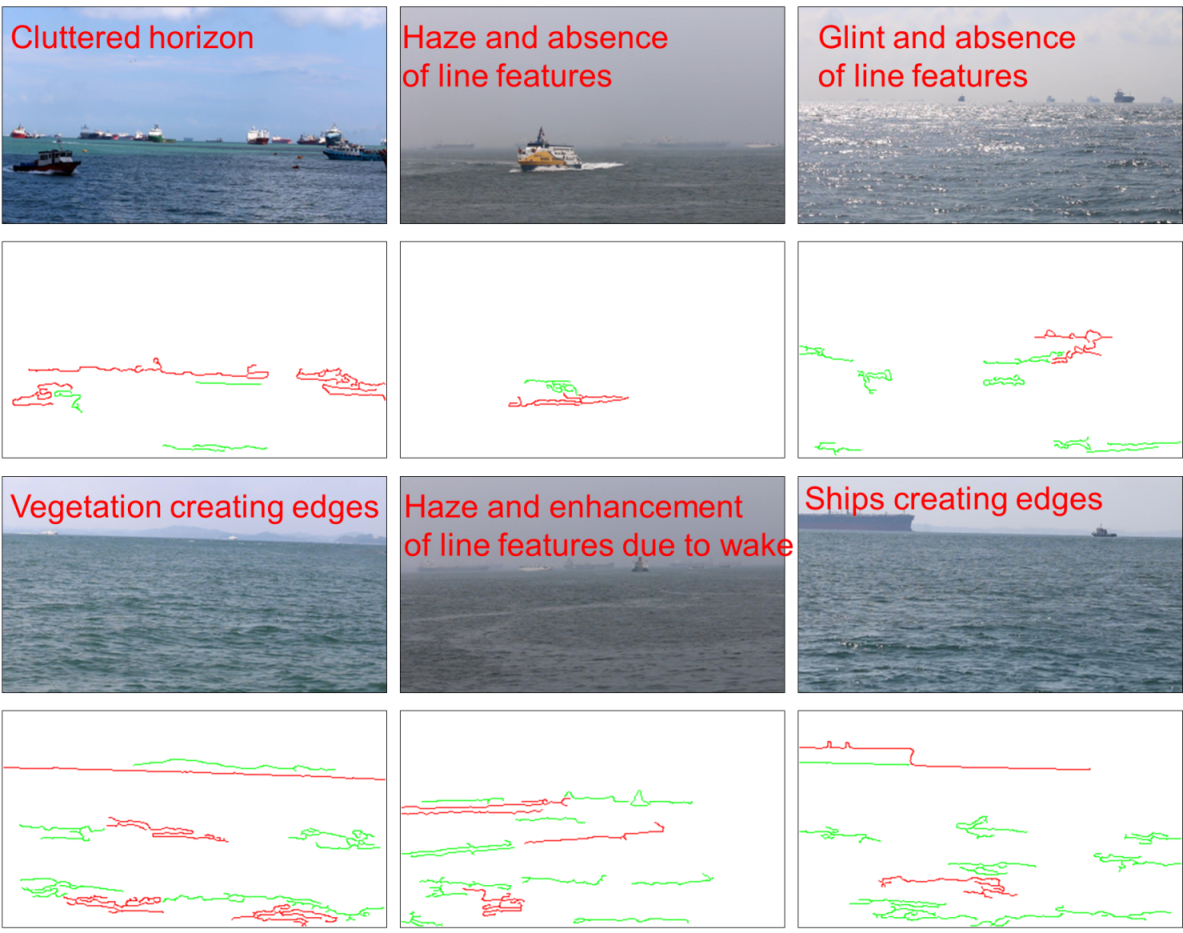}\\
    \vspace{-3mm}
  \caption{Challenging situations in which horizon detection is challenging}\label{fig:horiChallenging}
\end{figure}

Further, we note some frames from Singapore-Marine-dataset in Fig. \ref{fig:horiChallenging}, which are challenging due to reasons such as absence of line features of horizon, presence of competing line features (such as through ships and vegetation), adverse effects of conditions such as haze and glint, etc. For all these images, we show below them their edge maps where red edges are long edges and green edges are the edges of medium length. The dearth of line features representing horizon is evident in these edge maps. Also notable is that in conditions such as haze, the color distributions of sea and sky regions may be practically inseparable.

The statistical distance between sea and sky distributions may be increased by adding extra spectral channels \cite{prasad2015metrics} and abstract statistical distance metrics may be used through machine learning techniques \cite{fefilatyev2006horizon,prasad2016classification}. However, these approaches require sensor modification or their performance depends upon the diversity of the training dataset.

\section{Registration}\label{sec:registration}

Registration refers to the situation where different frames in a scene correspond to the same physical scene with matching world coordinates. In marine scenario, especially for sensors mounted on sea vessels and buoys, the unpredictable motion of the sensors often result in a complicated registration problem where even the consecutive frames are not registered and may have a large angular and positional shift, as noted in Table \ref{tab:Datasets}.

\begin{figure}
  \centering
  \includegraphics[width=\linewidth]{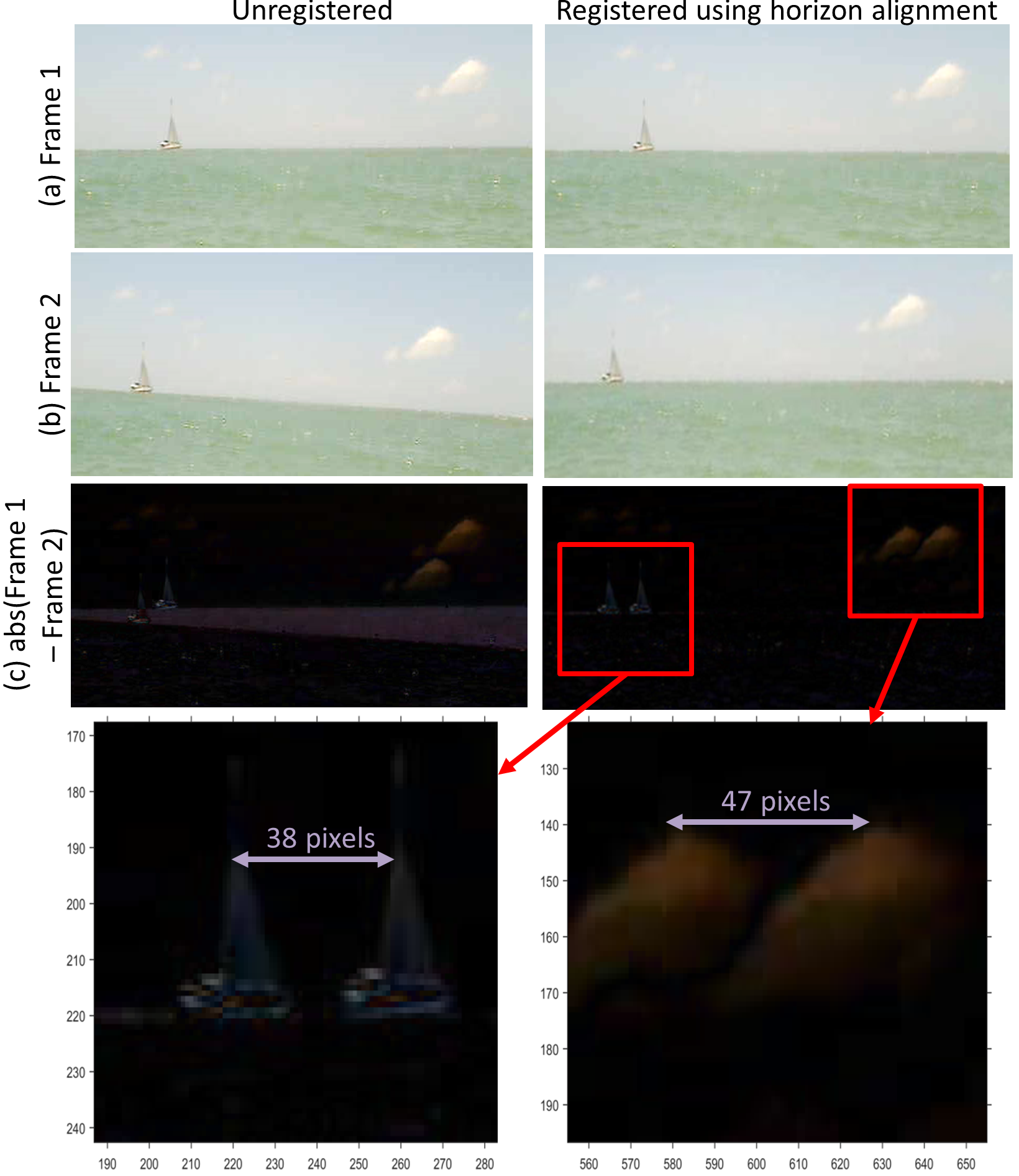}\\
  \vspace{-3mm}
  \caption{The top row shows the original consecutive frames and their difference from a video. Results of registration using horizon are shown in the second row. The third row shows two insets from difference image of the registered image.}\label{fig:registrationprobelm3}
\end{figure}

The angular difference between the two consecutive frames may have all the three angular components, viz. yaw, roll, and pitch. If the horizon is present, {roll and pitch can be significantly corrected for} since they result in the change of angle and position of the horizon, respectively. However, yaw cannot be corrected for. This is illustrated using two consecutive frames from a video in the buoy dataset are used in Fig. \ref{fig:registrationprobelm3}. It is seen that horizon based registration does reduce the differences (see middle row, 3rd image) but the zoom-ins shown in the bottom row clearly indicate that the boat and cloud have unequal horizontal difference between them. In this scenario, it is impossible to say if the cloud was stationary and the boat moved, or the boat was stationary and the cloud moved, or both of them moved.

\begin{figure}[t!]
  \centering
  \includegraphics[width=\linewidth]{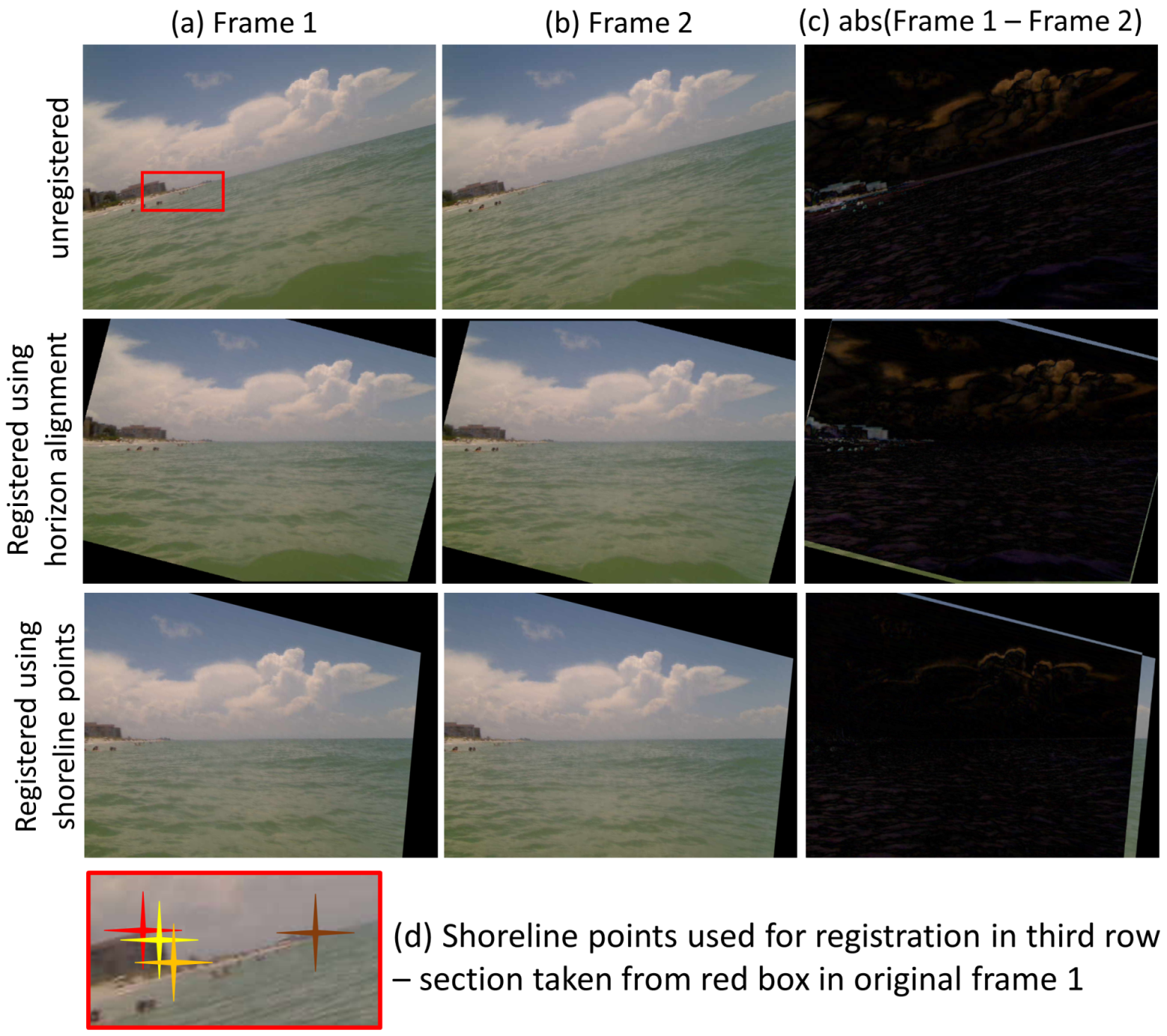}\\
  \vspace{-3mm}
  \caption{The top row shows the original consecutive frames and their difference from a video. Results of registration using horizon are shown in the second row. Registration results using just four fixed points on the shoreline are shown in the third row. The four points used for registration are shown in the last row.}\label{fig:registrationprobelm2}
\end{figure}

In order to correct for the yaw, we {need some additional features that allow the detection of the horizontal staggering between two consecutive frames}. The availability and possibility to detecting the stationary features is important for yaw correction. Buildings, landmarks, and terrain features may serve this purpose \cite{behringer1999registration}, if they are present in the scene. For example, we consider two consecutive frames in Fig. \ref{fig:registrationprobelm2} taken from another video in buoy dataset which does have stationary features. The result of registration using horizon only is shown in the middle row. However, using just a few manually selected stationary points on the shoreline, accuracy in registration is significantly enhanced, as seen in the third row.

Notably, although a ship may be stationary and can be easily detected, it is difficult to conclude whether the ship is stationary or not. Also, it is discussed in \cite{cao2008automatic} that the line features in a scene with moving vessels and absence of stationary cues may enable registration only if the vessels in the scene are not rotating. Thus, for a general maritime scenario, registration of frames is still a challenge. Strictly speaking, the best possible way of dealing with this scenario is the use of the ship's motion sensors and gyro sensors.

\begin{figure}
  \centering
  \includegraphics[width=\linewidth]{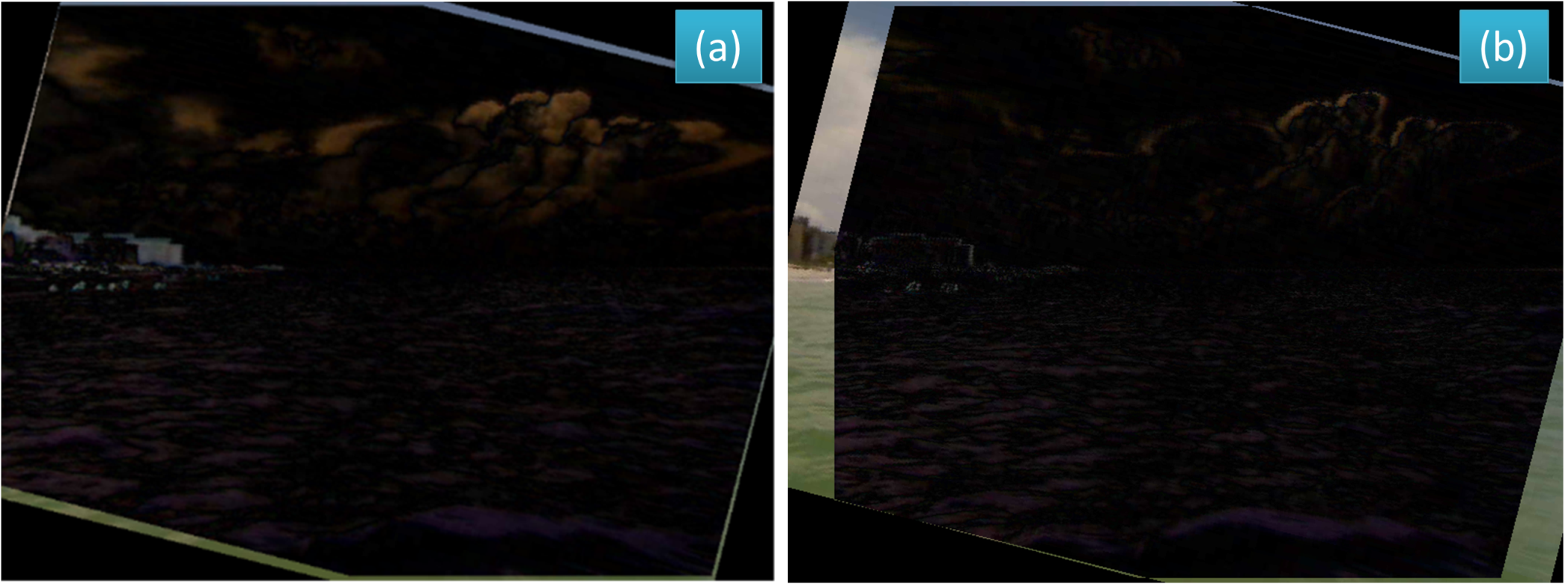}\\
  \vspace{-3mm}
  \caption{Registration using cross-correlation of strip around the horizon. (a) The difference image obtained by registration using horizon only, reproduced from Fig. \ref{fig:registrationprobelm2}. (b) The difference image after horizontal shift of 48 pixels, identified as the peak of the cross-correlation function.}\label{fig:Fefilatyev}
\end{figure}

Nevertheless, some help can be derived from texture-based features for registration across frames, assuming that the generalized shapes of texture boundaries might not change significantly over few consecutive frames \cite{criminisi2000shape}. Another related approach is used in \cite{fefilatyev2012algorithms} for registration, where a narrow horizontal strip is taken around the horizon in both the images and the shift at which the two images have maximum correlation is determined. This shift is used for registration. An example is shown in Fig. \ref{fig:Fefilatyev}. Optical flows may also be useful \cite{dusha2007attitude}, although at significant computation cost.

\section{Background subtraction}\label{sec:bckgnd}

There are several useful surveys on the topic of background suppression in video sequences \cite{elhabian2008moving}. Water background is more difficult than other stationary as well as dynamic backgrounds because of several reasons. One reason is that water background is continuously dynamic both in spatial and temporal dimensions due to waves, whereas the background subtraction methods typically address dynamic backgrounds that where dynamics are either spatially restricted (such as rustle of trees) or temporally restricted (such as a parked car). Second reason is that waves have a high spatio-temporal correlations \cite{ablavsky2003background} while the dynamic background subtraction methods implicitly infer high spatio-temporal correlations as patterned (i.e. non-random) movement of foreground objects. An associated difficulty in marine background detection is that the electro-optical sensor mounted on a mobile platform is subject to a lot of motion. Most background learning methods learn background by assuming that a pixel remains background or foreground for at least a certain period of time. Thus, background modelling depends upon the accuracy of registration, which is a challenging problem as discussed in the previous section. Third reason is that wakes, foams, and speckle in water are inferred as foreground by typical background detection method whereas they are background in the context of maritime object detection problem.

To illustrate the need for new algorithms addressing maritime background, we applied the 34 algorithms that participated in the change detection competition \cite{sobral2014comprehensive}. This competition was conducted in 2014 as a part of a change detection workshop at a prestigious computer vision conference \cite{wang2014cdnet}. It used a dataset of 51 videos comprising of about 140,000 frames separated into 11 categories of background challenges such as dynamic background, camera jitter, intermittent object motion, shadows, infrared videos, snow, storm, fog, low frame rate, night videos, videos from pan-tilt-zoom camera, and air air turbulence. Since the dataset addressed several background challenges encountered in maritime videos as well and the submitted algorithms represented the state-of-the-art for these challenges, we tested their performance on Singapore-Marine-dataset.

\begin{figure*}
  \centering
  \includegraphics[width=0.95\linewidth]{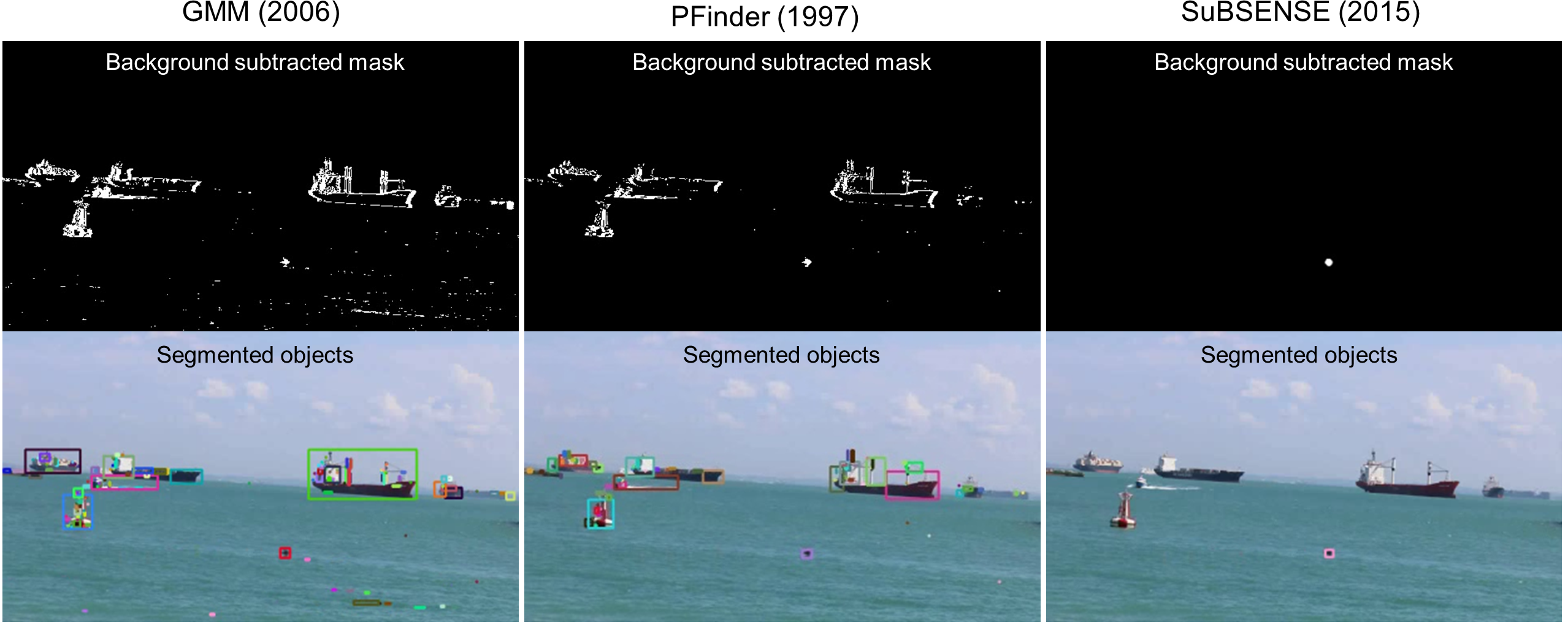}\\
  \caption{Results of three methods from change detection competition that perform the best or fastest.}\label{fig:background}
\end{figure*}

Here, we show in Fig. \ref{fig:background} the result of three methods for one frame of a video from on-shore Singapore-Marine dataset. The three methods are Gaussian mixture model (GMM) \cite{zivkovic2006efficient}, which models background's color distribution as mixture of Gaussian distributions, Gaussian background model of PFinder \cite{wren1997pfinder}, which models the intensity at each background pixel as a single Gaussian function and then clusters these Gaussian functions as representing the background, and the self-balancing sensitivity segmenter (SuBSENSE) \cite{st2015subsense}, which uses local binary similarity patterns at pixel levels for modeling background. It is seen that these methods are ineffective through producing false positives in the water region or through producing false negatives while suppressing water background.

\section{Foreground object detection}\label{sec:object}\label{sec:foreground}

Even with proper dynamic background subtraction, such that wakes, foams, clouds, etc. are suppressed, it is notable that further foreground segmentation can result in detection of mobile objects only. However, as noted in Table \ref{tab:Datasets}, there are several stationary objects as well in the videos. In Table \ref{tab:Datasets}, the ground truth for stationary and dynamic objects have been generated for each video manually by independent volunteers. The segmented background has to be further analysed for detecting the static foreground objects. Since the general dynamic background subtraction and foreground tracking problems do not require the detection of static objects, no integrated approaches exist that can simultaneously detect the stationary and mobile foreground objects. This is an open challenge for the maritime scenario. Research for the problem of object detection in images may be applied for detection of objects in individual images, thus catering for both static and mobile objects. However, the complicated maritime environment with potential of occlusion, orientation, scale, and variety of objects make it computationally challenging \cite{bloisi2011automatic}. Further, complicated motion patterns imply that frame to frame matching of objects for tracking is challenging if detection is performed independently for each frame.

\section{Weather and illumination conditions}\label{sec:weather}

A maritime scene is subjected to a vast variety of weather and illumination conditions such as bright sunlight, twilight conditions, night, haze, rain, fog, etc. Further, the solar angles induce different speckle and glint conditions in the water. Tides also influence the dynamicity of water. The situations that affect the visibility influence the contrast, statistical distribution of sea and water, and visibility of far located objects. Effects such as speckle and glint create non-uniform background statistics which need extremely complicated modelling such that foreground is not detected as the background and vice versa. Also, the color gamuts for illumination conditions such as night (dominantly dark), sunset (dominantly yellow and red), and bright daylight (dominantly blue), and hazy conditions (dominantly gray) also vary significantly. As a consequence, the suitable methods and models for one weather and illumination condition is not effective for other conditions. Seamless selection of approaches and transition between one approach to another with varying conditions is important for making maritime processing practically useful.

\section{Conclusion}
As discussed above, maritime video processing problem poses challenges that are absent or less severe in other video processing applications. It needs unique solutions that address these challenges. It also needs algorithms with better adaptability to the various conditions encountered in maritime scenario. Thus, the field is rich with possibilities of innovation in maritime video processing technology. We hope that the discussion here motivates the researchers to pursue maritime video processing challenges with enthusiasm and vigour.


\end{document}